# Characterizing and Predicting Wildfire Evacuation Behavior: A Dual-Stage ML Approach

Sazzad Bin Bashar Polock
*Ingram School of Engineering*
*Texas State University*
San Marcos, TX, USA
sazzadpolock8@gmail.com

Anandi Dutta, Ph.D.
*Ingram School of Engineering*
*Texas State University*
San Marcos, TX, USA
anandi.dutta@txstate.edu

Subasish Das
*Ingram School of Engineering*
*Texas State University*
San Marcos, TX, USA
subasish@txstate.edu

*Abstract*— Wildfire evacuation behavior is highly variable and influenced by complex interactions among household resources, preparedness, and situational cues. Using a large-scale MTurk survey of residents in California, Colorado, and Oregon, this study integrates unsupervised and supervised machine learning methods to uncover latent behavioral typologies and predict key evacuation outcomes. Multiple Correspondence Analysis, K-Modes clustering, and Latent Class Analysis reveal consistent subgroups differentiated by vehicle access, disaster planning, technological resources, pet ownership, and residential stability. Complementary supervised models show that transportation mode can be predicted with high reliability from household characteristics, whereas evacuation timing remains difficult to classify due to its dependence on dynamic, real-time fire conditions. These findings advance data-driven understanding of wildfire evacuation behavior and demonstrate how machine learning can support targeted preparedness strategies, resource allocation, and equitable emergency planning.

## I. INTRODUCTION

Wildfires have grown increasingly severe across the western United States, particularly in California, Colorado, and Oregon, where expanding wildland–urban interface development and climate-driven factors have heightened fire risk. In contrast to other natural hazards, wildfires spread rapidly and unpredictably, presenting distinct evacuation challenges that demand immediate decision-making under high-risk conditions [1]. Evacuation behavior during wildfire events is highly complex and heterogeneous, shaped by a range of factors including demographic characteristics, risk perception, prior experience, and access to resources [2], [3]. Research grounded in the Protective Action Decision Model (PADM) has identified that age, gender, income, and household composition significantly predict evacuation responses, with females more likely to evacuate than males and older adults showing greater reluctance to leave [4], [5]. The presence of dependents, such as children, elderly family members, or pets, often complicates and delays evacuation decisions [4], [6]. Additionally, technological access to smartphones and GPS systems can enhance evacuation capacity, though disparities may contribute to unequal outcomes [7], [8].

Despite growing interest in wildfire evacuation behavior, systematic studies grounded in large-scale survey data remain limited [2], [3]. Most research relies on post-disaster interviews or simulation models that may not fully capture real-time decision-making complexity [1]. The PADM provides a theoretical framework for understanding how individuals process information and make protective action decisions during hazard events [5], [9], [10]. However, empirical applications of PADM to wildfire contexts remain relatively scarce, particularly in the United States [2], [11]. Recent methodological advances include online platforms like Amazon Mechanical Turk (MTurk) for behavioral data collection. Despite concerns about sample representativeness, it can provide valuable insights into evacuation behavior patterns at scale when combined with rigorous quality control procedures [12]. Studies utilizing MTurk have demonstrated comparable data quality to traditional survey methods when appropriate screening and validation measures are implemented. In this project, we apply both supervised and unsupervised machine learning techniques to a large-scale wildfire evacuation survey collected from households in California, Colorado, and Oregon. Our goal is to identify underlying patterns in evacuation behavior and to build predictive models capable of classifying key decisions such as evacuation timing and travel mode.

## II. LITERATURE REVIEW

Evacuation behavior during wildfire events displays dynamics that distinguish it from other hazard contexts, such as hurricanes or floods. Unlike slow-onset disasters, wildfires are marked by rapid ignition, unpredictable spread driven by shifting wind conditions, and high threat visibility, all of which shape resident decision-making under severe time constraints. While simulation studies often assume optimal behavior, actual responses are frequently more variable, with many households delaying evacuation due to last-minute preparations, uncertainty regarding fire proximity, or reliance on unofficial information sources [13]. Such delays can critically impact clearance times and safety outcomes [13], [14]. Evacuation behavior is influenced by both individual and situational factors, including prior experience, trust in official warnings, and perceived capacity to defend property [15], [16], [17]. Moreover, some residents choose to remain and actively defend their homes, particularly in areas where evacuation orders are advisory rather than mandatory [18], [19]. This behavior is influenced by cultural norms, perceived self-efficacy, and infrastructural constraints, such as limited or single-access road networks. Extending this perspective, residents' stated intentions across evacuation options, including evacuation, staying to defend, or sheltering in place, indicate that these choices are associated with property-level mitigation measures and demographic characteristics [20]. In addition, full-time rural residents exhibited a greater tendency to remain and defend their properties, particularly when they perceived their homes to be sufficiently prepared to withstand wildfire exposure without external firefighting assistance [21], [22]. Complementing these findings, discrete choice modeling of a wildfire evacuation in Israel, illustrated that travel behavior under wildfire threat diverges significantly from routine patterns [23].

A broad set of household- and individual-level factors shape evacuation decision-making during wildfire events. Socio-demographic characteristics, including age, income, gender, and household size, are commonly linked to differences in evacuation behavior [24], [25], [26], [27]. For instance, older adults and males have been found to be less likely to evacuate compared to younger individuals and females, reflecting differences in risk perception and willingness to act [2], [3], [28], [29]. Furthermore, access to key resources such as private vehicles, financial means, and temporary shelters significantly affects the feasibility of evacuation, especially under time constraints [30]. The presence of dependents including children, elderly family members, or pets has also been identified as a determinant that complicates the evacuation decision, often increasing hesitation or delay [2]. Furthermore, subjective risk perception plays a critical role in evacuation decision-making. Individuals who perceive wildfire threats as imminent or severe are more likely to evacuate early, whereas those who believe they can defend their property or who underestimate the hazard often delay or decline evacuation [2]. Additionally, the decision-making process is further shaped by attitudes toward personal safety and property protection, with some individuals prioritizing the latter due to emotional attachment or perceived self-efficacy in managing the threat [30]. Moreover, access to technology such as smartphones and GPS applications has emerged as another influential factor, enabling real-time information retrieval, route optimization, and situational awareness, all of which enhance household capacity to respond quickly and adaptively [30], [31], [32].

Structured surveys are a critical tool in disaster and evacuation research, enabling the systematic collection of individual-level behavioral data across diverse populations [33], [34]. These instruments are particularly effective in capturing real-time decision-making processes, perceptions of risk, and patterns of behavioral heterogeneity during emergency scenarios. Additionally, online platforms such as Amazon MTurk have gained traction for their scalability and cost-efficiency in behavioral studies [35], [36]. However, concerns remain regarding sample representativeness, as MTurk respondents often skew younger, more educated, and technologically savvy than the general population, potentially limiting generalizability [37]. Nonetheless, researchers have employed rigorous data quality controls to enhance validity, including attention checks, consistency filters, and question randomization [38]. When applied appropriately, survey-based approaches yield valuable insights into the dynamics of evacuation behavior, particularly when combined with complementary data sources such as social media, app usage, and geospatial tracking [39]. For example, surveys conducted after the Fort McMurray wildfire demonstrated the utility of crowdsourcing platforms for evaluating app feature relevance, linking public needs with mobile software design during crises [40], [41]. Thus, despite their limitations, structured surveys remain essential for uncovering the social and cognitive dimensions of evacuation choices at scale.

Prior studies on wildfire evacuation have mainly used statistical analyses, agent-based simulations, or discrete choice models to study evacuation decision-making. These traditional approaches often assume linear relationships among factors or allocate individuals to predefined behavioral categories, which can limit their ability to capture the complexity and interdependence inherent to real-world evacuation behavior. While valuable, such models can struggle to handle high-dimensional data and dynamic behavioral patterns found during actual evacuations [42], [43], [44]. More recently, machine learning has shown promise in disaster research, including wildfire contexts, by enabling accurate prediction of evacuation compliance and risk classification, and the identification of patterns in large datasets. Examples include supervised ML models predicting evacuation compliance and unsupervised models revealing unexpected behavioral clusters. However, few studies have integrated unsupervised behavioral clustering (such as Latent Class Analysis, Multiple Correspondence Analysis, and K-modes cluster) with supervised learning (like Random Forest or XGBoost) to analyze large-scale, self-reported household survey data in wildfire evacuations [45], [46]. This study addresses this gap by integrating advanced clustering methods with classification models to both discover latent evacuation response profiles and predict essential evacuation outcomes using survey data. Such an approach advances methodological rigor in wildfire evacuation research and provides actionable recommendations for segment-specific planning and communication by emergency management [47], [48].

III. DATASET DESCRIPTION

The dataset used for this analysis was obtained from an online survey distributed through Amazon Mechanical Turk (MTurk) in April–May 2022. Participants were residents of California, Colorado, and Oregon who were living in wildfire-prone areas. A total of 1,312 survey responses were initially collected. To ensure data quality, a three-stage screening procedure was applied, including attention-check questions, repeated-question consistency verification, and a logical household composition rule. Responses were flagged if reported household counts (e.g., children or elderly) exceeded total household size, contained negative values, or were missing for key variables. Because such inconsistencies could not be reliably resolved, flagged responses were removed rather than corrected or imputed. After applying these quality control filters, 853 valid responses were retained for subsequent analysis. Despite these quality control procedures, certain limitations related to the survey recruitment approach should be acknowledged. While rigorous screening procedures were employed, the use of Amazon Mechanical Turk introduces known sampling limitations. MTurk respondents tend to be younger, more educated, and more technologically connected than the general population, which may overrepresent households with higher access to smartphones, navigation tools, and online information. Consequently, technology-dependent evacuation behaviors observed in this study may not fully reflect the experiences of elderly, low-income, or digitally marginalized populations. Accordingly, the findings are intended to identify behavioral patterns and typologies rather than estimate population-level prevalence or representativeness. Future studies should complement online surveys with community-based or offline data collection to improve generalizability. The survey captures a wide range of variables relevant to evacuation decision-making, including sociodemographic characteristics (e.g., age, income, household size), mobility factors (e.g., number of vehicles, access to GPS or smartphones), risk perception and experience (e.g., past evacuations, length of residence), and key behavioral outcomes such as evacuation timing, evacuation trigger, transportation mode, route choice, and destination selection. Additional contextual factors such as surrounding infrastructure, pet or livestock ownership, and access to formal or informal shelters further enrich the dataset and broaden its applicability for modeling evacuation behavior. As part of this study, we also conducted an exploratory analysis of all major variables to understand distributional patterns, missingness, and inter-variable relationships.

## IV. METHODOLOGY

This work uses a hybrid machine-learning approach combining both unsupervised and supervised techniques to analyze wildfire evacuation behavior. First, the survey data is cleaned and transformed into machine-readable form, with categorical variables encoded and key outcome variables, such as early versus late evacuation and transportation mode, clearly defined. Unsupervised methods, including Multiple Correspondence Analysis (MCA), K-Modes clustering, and Latent Class Analysis (LCA), are then applied to identify natural groups of evacuees based on shared characteristics. These clusters help reveal underlying behavioral patterns related to preparedness, mobility, and decision timing. Next, supervised models such as Logistic Regression, Random Forest, and XGBoost are used to predict two important outcomes: whether a respondent evacuates early or late and whether they use private or non-private transportation. Models are trained on a 75/25 train–test split and evaluated using accuracy, F1-score, Kappa, and ROC-AUC. Feature importance and explainability tools help interpret the key factors influencing predictions. Finally, cluster labels from the unsupervised stage may be added as additional features to test whether they improve predictive performance. This integrated approach provides both descriptive insights and predictive modeling of wildfire evacuation behavior. In Figure 1, this dual-stage ML approach is presented.

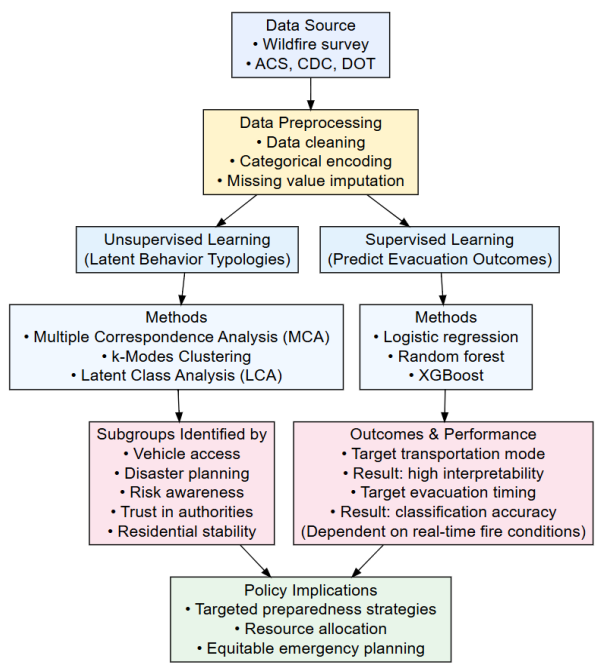

Figure. 1. A dual-stage ML approach

## V. RESULT AND DISCUSSION

### A. Unsupervised Clustering Analysis

Unsupervised learning revealed a multilayered structure within the wildfire evacuation dataset, offering valuable insight into how preparedness, mobility resources, household composition, and decision-making tendencies vary across the surveyed population. The MCA scatterplot, Figure 2, showed that respondents occupy a broad and continuous behavioral space rather than forming sharply separated groupings, indicating that wildfire evacuation behavior is best understood as a spectrum of tendencies rather than discrete categories.

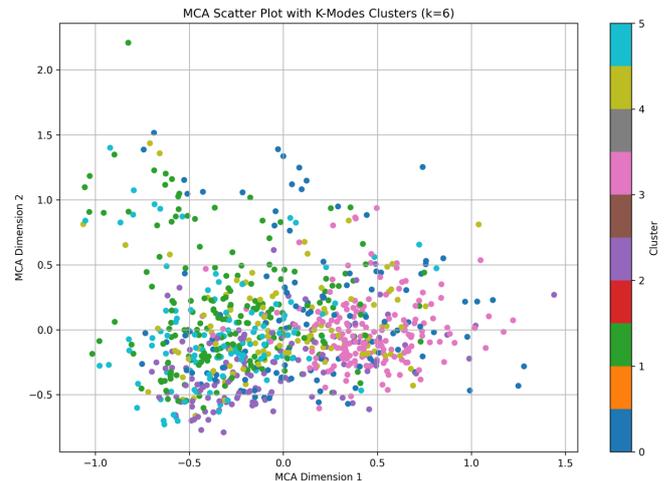

Figure. 2. MCA scatterplot showing respondent distribution and K-Modes clustering assignment (K=6)

This pattern was further supported by the MCA category map in Figure 3, which displayed the spatial relationships among individual response categories and revealed that highly common evacuation-related characteristics, such as traveling by private vehicle, selecting family or friends as an evacuation destination, or lacking a written disaster plan, clustered near the center of the space, reflecting their widespread use across respondents. Because these behaviors appeared frequently in the dataset, their corresponding category points were drawn toward the dense central region where many overlapping patterns of household decision-making converge. In contrast, less common or more specialized behaviors, such as evacuating on foot, relying on non-car modes like bicycles or motorcycles, or choosing destination types outside typical residential or hospitality settings, were positioned toward the periphery of the map. Their separation from the central mass indicates that these response patterns are not only rare but also associated with more distinct combinations of household characteristics and evacuation contexts.

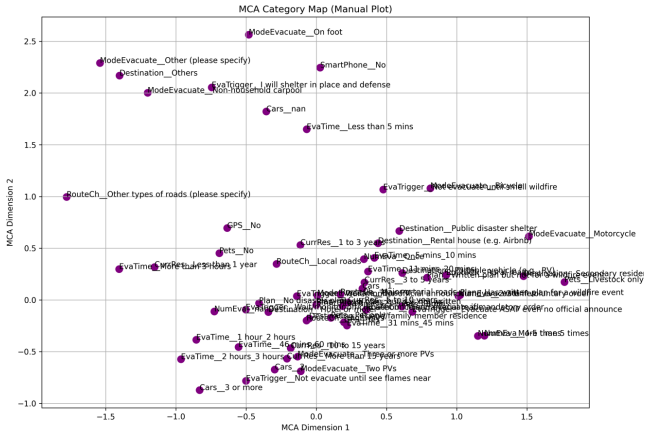

Figure. 3. MCA plot of evacuation-related categories positioned along the first two latent dimensions.

Building on the MCA representation (Figure 4), the six-cluster K-Modes solution produced a set of moderately distinct segments that, while not sharply separated in space, nonetheless captured consistent differences in household resources and behavioral patterns.

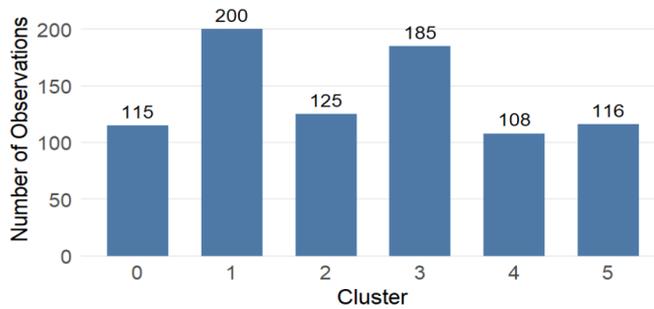

Figure. 4. Distribution of households across the six behavioral clusters identified by the K-modes algorithm (k = 6).

Based on the behavioral profiles derived from the K-modes clustering, the six clusters highlighted meaningful population subgroups. Cluster 0, for example, consisted mostly of households with limited access to private vehicles, minimal or no planning, and lower levels of technological support (such as smartphones or GPS), suggesting that these respondents may face practical barriers to rapid evacuation. In contrast, Cluster 1 aligned more closely with stable, long-term residents possessing multiple vehicles, where private-vehicle use and family-based evacuation destinations were common, indicating a more confident and predictable evacuation process. Cluster 2 tends to represent short-term renters or recently relocated individuals, many of whom reported constrained mobility or weaker disaster-preparedness practices, which may limit their flexibility during an emergency. Cluster 3 stood out as the best-prepared subgroup, composed of households with multiple vehicles, widespread smartphone and GPS access, and a higher likelihood of having a written evacuation plan; this cluster often demonstrated faster and more deliberate evacuation tendencies. Cluster 4 captured households with multiple pets or livestock, whose evacuation decisions involve additional logistical coordination and may therefore be influenced by animal-related constraints. Finally, Cluster 5 emerged as a smaller, more heterogeneous group with inconsistent or ambiguous response patterns, suggesting that some households may not fit cleanly into any one behavioral type or may have less structured evacuation decision processes.

To further validate these typologies, a complementary Latent Class Analysis was conducted, producing a six-class solution whose class-size distribution and posterior probability heatmap demonstrated that similar behavioral dimensions emerge even when using a probabilistic clustering approach (Figures 5 and 6).

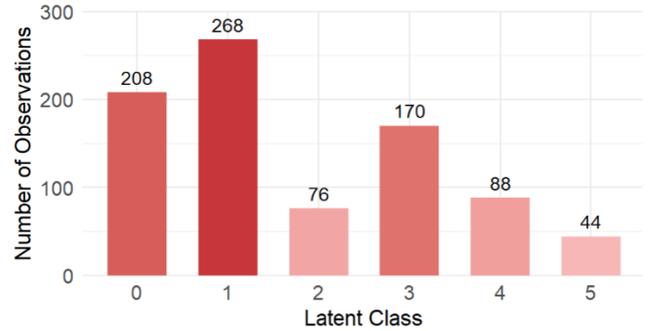

Figure. 5. Household distribution across the six latent classes from the LCA model, showing the relative size of each behavioral subgroup.

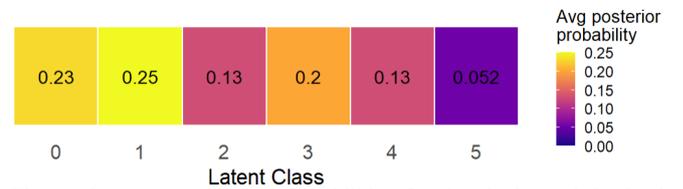

Figure. 6. Average posterior probabilities for the six latent behavioral classes.

The LCA results produced a set of latent class profiles that closely mirrored the K-Modes behavioral groupings: the two largest classes aligned with the well-prepared, vehicle-reliant households seen in K-Modes Clusters 1 and 3, while several smaller classes reflected the resource-limited, lower-planning, or uncertainty-driven patterns characteristic of Clusters 0, 2, and 5. Although the posterior probabilities were partially diffuse, indicating that some respondents did not fit cleanly into a single class, the overall similarity between the LCA profiles and the K-Modes clusters suggests a stable underlying structure in how households organize their wildfire evacuation decisions. Taken together, the MCA visualizations, K-Modes clusters, and LCA profiles reveal a consistent continuum of behavioral typologies shaped by differences in vehicle availability, preparedness planning, pet and livestock responsibilities, technological access, and residential stability. These converging findings emphasize that wildfire evacuation behavior is not homogeneous but reflects a diverse set of strategies influenced by each household's constraints and capacities, insights that hold clear implications for designing targeted preparedness messaging and equitable evacuation support systems.

### B. Supervised Learning Results

Supervised machine learning models were applied to evaluate how effectively household characteristics can predict two key evacuation outcomes: departure timing and transportation mode. For this analysis, early evacuation was defined as leaving within roughly 5–20 minutes, whereas late evacuation encompassed longer response windows such as departing after 30 minutes, waiting for official announcements, or monitoring conditions before leaving. This early–late distinction is consistent with operational thresholds used in wildfire evacuation research, where rapid departure can be critical in fast-moving fire scenarios. Across both tasks, the predictive performance reveals important insights into the nature of wildfire decision-making. The first task (Table 1), early versus late evacuation, proved considerably more difficult to model. Logistic Regression achieved the highest performance, with an accuracy of 0.62, weighted F1 score of 0.62, and ROC-AUC of 0.61. Random Forest performed slightly lower, with an accuracy of 0.58 and ROC-AUC of approximately 0.62, indicating modest but consistent predictive ability. XGBoost showed the weakest performance with an accuracy of 0.53. These modest values indicate that, although the models captured some of the underlying patterns, a substantial portion of variation in evacuation timing arises from unobserved factors not represented in survey features. Timing decisions are heavily influenced by situational dynamics, such as real-time smoke cues, fire proximity, community alerts, or personal risk tolerance, which cannot be fully inferred from static household attributes. As a result, the relatively low Cohen's Kappa values suggest limited agreement beyond chance, reinforcing that evacuation timing is inherently noisy and difficult to classify using structured data alone.

TABLE 1. PREDICTIVE PERFORMANCE OF MODELS FOR CLASSIFYING HOUSEHOLDS INTO EARLY VERSUS LATE EVACUEES.

| Model | Accuracy | F1 | ROC-AOC | Cohen's K |
|---|---|---|---|---|
| Logistic Regression | 0.62 | 0.62 | 0.61 | 0.24 |
| Random Forest | 0.58 | 0.58 | 0.62 | 0.17 |
| XGBoost | 0.53 | 0.53 | 0.55 | 0.07 |

In contrast, the second task (Table 2), predicting evacuation mode, yielded much stronger and more reliable outcomes across all models. Logistic Regression reached an accuracy of 0.89, Random Forest produced 0.87, and XGBoost performed similarly at 0.86, with consistently high weighted F1 scores. These results indicate that transportation mode is far more structurally determined by household characteristics than timing. High performance, particularly for private-vehicle users, reflects the prevalence and stability of this choice, which is closely tied to vehicle ownership, planned destinations, and household preparedness practices. Even though minority classes such as non-car evacuees were more challenging due to their smaller sample size, the overall predictive strength demonstrates that mode choice can be inferred with substantial confidence from survey-reported variables.

TABLE 2. PERFORMANCE OF MODELS FOR EVACUATION MODE PREDICTION (3-CLASS).

| Model | Accuracy | F1 | Cohen's K |
|---|---|---|---|
| Logistic Regression | 0.89 | 0.87 | 0.15 |
| Random Forest | 0.87 | 0.87 | 0.22 |
| XGBoost | 0.86 | 0.86 | 0.17 |

The findings suggest that while timing prediction remains challenging due to real-time behavioral variability, transportation mode can be forecasted with high reliability using static household features. This distinction is meaningful from a planning perspective. Accurate mode prediction supports transportation load forecasting, shelter allocation, and resource distribution. In contrast, the weaker predictability of timing highlights the need for dynamic warning systems, continuous monitoring, and real-time behavioral modeling beyond survey attributes.

## VI. CONCLUSION

Wildfire evacuations remain difficult to manage because households differ widely in when they choose to leave and how they travel. This study addressed that challenge by applying a combined machine learning framework to identify hidden behavioral typologies and evaluate the predictability of key evacuation outcomes. Using Multiple Correspondence Analysis, K-Modes clustering, and Latent Class Analysis, the study uncovered consistent behavioral groups shaped by vehicle access, preparedness, pet ownership, and household stability. These unsupervised findings were complemented by supervised models, Logistic Regression, Random Forest, and XGBoost, to assess whether household characteristics can reliably predict evacuation timing and transportation mode.

The results show that transportation mode is highly predictable, with all models achieving strong accuracy, while evacuation timing remains difficult to model due to real-time situational influences not captured in survey data. These patterns offer actionable insights for emergency planners. Predictable mode choice can directly inform traffic management, roadway capacity planning, and targeted assistance for low-mobility or non-vehicle households. Conversely, the limited predictability of departure timing underscores the need for improved alert systems, dynamic communication strategies, and real-time situational intelligence rather than static demographic-based predictions. Together, the findings highlight that while household structure strongly shapes how people evacuate, when they choose to leave requires flexible, adaptive policy interventions.

## VII. FUTURE WORK

Future research should integrate real-time hazard indicators, such as fire spread models, smoke visibility data, and warning system logs, to improve the prediction of evacuation timing, which remained difficult to capture using static survey features. Incorporating mobility traces, sensor-based risk alerts, or longitudinal behavioral data may also strengthen the modeling of dynamic decision processes. Additionally, expanding the framework to include equity-focused analyses could help identify communities that face disproportionate evacuation barriers. Applying this hybrid machine learning approach to other hazard contexts, such as hurricanes or floods, would further test its generalizability and practical value.